\documentclass{article}
\usepackage{spconf,graphicx}
\usepackage{graphicx}
\usepackage[fleqn]{amsmath}
\usepackage{amssymb}
\usepackage{url}
\usepackage{float}
\usepackage{dsfont}
\usepackage{algorithm}% http://ctan.org/pkg/algorithms
\usepackage{algpseudocode}% http://ctan.org/pkg/algorithmicx
\usepackage{color}
\usepackage{multirow}
\usepackage{bbm}
\usepackage{bm}
\usepackage{mathtools}
\usepackage{hyperref}

\usepackage[table,xcdraw]{xcolor}

\DeclareMathOperator*{\argmin}{argmin}

\def\mbf{\mathbf}

\def\L{{\cal L}}
\def\R {\mathbb{R}}

\title{SPEECH-TO-SPEECH TRANSLATION BETWEEN UNTRANSCRIBED UNKNOWN LANGUAGES}
\name{Andros Tjandra$^{1}$, Sakriani Sakti$^{1,2}$, Satoshi Nakamura$^{1,2}$\thanks{}}
\address{	$^1$Nara Institute of Science and Technology, Japan\\
	$^2$RIKEN, Center for Advanced Intelligence Project AIP, Japan\\ \texttt{\{andros.tjandra.ai6,ssakti,s-nakamura\}@is.naist.jp}}
\begin{document}
	\ninept
	\maketitle
	\begin{abstract}
		In this paper, we explore a method for training speech-to-speech translation tasks without any transcription or linguistic supervision. Our proposed method consists of two steps: First, we train and generate discrete representation with unsupervised term discovery with a discrete quantized autoencoder. Second, we train a sequence-to-sequence model that directly maps the source language speech to the target language’s discrete representation. Our proposed method can directly generate target speech without any auxiliary or pre-training steps with a source or target transcription. To the best of our knowledge, this is the first work that performed pure speech-to-speech translation between untranscribed unknown languages.
	\end{abstract}

	\section{Introduction}
	\label{sec:intro}
	
	Information exchanges among different countries continue to increase. International travelers for tourism, emigration, or foreign study are becoming increasingly diverse, heightening the need for devising a means to offer effective interaction among people who speak different languages. Since automatic spoken-to-speech translation (S2ST) provides an opportunity for people to communicate in their own languages, it significantly overcomes language barriers and closes cross-cultural gaps.
	
	Many researchers have been developing a S2ST system over the past several decades. A traditional approach in S2ST systems requires effort to construct several components, including automatic speech recognition (ASR), machine translation (MT), and text-to-speech (TTS) synthesis, all of which are trained and tuned independently. Given speech input, ASR processes and transforms speech into text in the source language, MT transforms the source language text to corresponding text in the target language, and finally TTS generates speech from the text in the target language. Significant progress has been made and various commercial speech translation systems are already available for several language pairs. However, more than 6000 languages, spoken by 350 million people, have not been covered yet. Critically, over half of the world's languages actually have no written form; they are only spoken.  
	
	Recently, end-to-end deep learning frameworks have shown impressive performances on many sequence-related tasks, such as ASR, MT, and TTS \cite{chorowski2015attentionasr,bahdanau2014nmt,wang2017tacotron}. Their architecture commonly uses an attentional-based encoder-decoder mechanism, which allows the model to learn the alignments between the source and the target sequence, that can perform end-to-end mapping tasks of different modalities. Many complicated hand-engineered models can also be simplified by letting neural networks find their way to map from input to output spaces. Thus, the approach provides the possibility of learning a direct mapping between the variable-length of the source and the target sequences that are often not known a priori. Several works extended the sequence-to-sequence model’s coverage by directly performing end-to-end speech translation using only a single neural network architecture instead of separately focusing on its components (ASR, MT, and TTS).
	
	Although the first feasibility was shown by Duong et al. \cite{duong2016sptranslation}, they focused on the alignment between the speech in the source language and the text in the target language because their speech-to-word model did not yield any useful output. The first full-fledged end-to-end attentional-based speech-to-text translation system was successfully performed by B$\acute{e}$rard et al. on a small French-English synthetic corpus \cite{berard2016proof}. But their performance was only compared with statistical MT systems. Weiss et al. \cite{weiss2017direct} demonstrated that end-to-end speech-to-text models on Spanish-English language pairs outperformed neural cascade models. Kano et al. then proved that this approach is possible for distant language pairs such as Japanese-to-English translation \cite{kano2017structured}. Similar to the model by Weiss et al. \cite{weiss2017direct}, although it does not explicitly transcribe the speech into text in the source language, it also doesn’t require supervision from the groundtruth of the source language transcription during training. However, most of these works remain limited to speech-to-text translation and require text transcription in the target language.
	
	Recently, Jia et al. \cite{jia2019direct} proposed the deep learning model that is trained end-to-end, which learns to map speech spectrograms into target spectrograms in another language that corresponds to the translated content (in a same or different canonical voice). Unfortunately, since training without auxiliary losses leads to extremely poor performance, they provided a solution by integrating auxiliary decoder networks to predict phoneme sequences that  correspond to the source and/or target speech. Despite much progress in direct speech translation research, no completely direct speech-to-speech translation has been achieved without any text transcription in source and target languages, during training, has not been achieved yet. Therefore, it remains difficult to scale-up the existing approach to unknown languages without written forms or transcription data available.
	
	On the other hand, there has been a project that held by speech community to push toward developing unsupervised, data-driven systems that are less reliant on linguistic expertise. Zero resource modeling is an approach where completely unsupervised techniques can learn the elements of a language’s speech hierarchy solely from untranscribed audio data. This means that only spoken audio data are
	available in a specific language, but transcriptions, annotations, and prior knowledge for it are all unavailable. The “Zero Resource Speech Challenge” s eries \cite{versteegh2015zero,dunbar2017zero,dunbar2019zero} was constructed to progress incrementally toward  a system that learns an end-to-end spoken dialog (SD) system in an unknown language from scratch just using information available to language learning infants. The ZeroSpeech 2019 \cite{dunbar2019zero} challenge confronts the problem of constructing a speech synthesizer without any text or phonetic labels: TTS without T . It is a continuation of the subword unit discovery track of ZeroSpeech 2015 and 2017 \cite{versteegh2015zero, dunbar2017zero}. 19 systems were submitted, but few studies proposed end-to-end frameworks \cite{liu2019zero,cho2019zero,andros2019zero}. Among these proposed systems, the vector quantized variational autoencoder (VQ-VAE) approach provides a better performance of naturalness based on mean opinion score (MOS) on the generated speech and character error rate after human transcription of the speech synthesis. Further details of the results are available: \href{www.zerospeech.com/2019/results.html}{\url{www.zerospeech.com/2019/results.html}}.
	
	In this paper, we take a step beyond the task of the current ZeroSpeech 2019 and propose a method for training speech to speech translation tasks without any transcription or linguistic supervision. Instead of only discovering subword units and synthesizing them within a certain language, our approach discovers subword units that are directly translated to another language. Our proposed method consists of two steps: (1) we train and generate discrete representation with unsupervised term discovery, which is also based on a discrete quantized autoencoder; (2) we train a sequence-to-sequence model to directly map the source language speech to the target language discrete representation. Our proposed method can directly generate target speech without any auxiliary or pre-training steps with source or target transcription. To the best of our knowledge, this is the first work that performed pure speech-to-speech translation between untranscribed unknown languages.

	\section{Unsupervised Unit Discovery with VQ-VAE} 
	\label{sec:vqvae}
	A speech signal can be disentangled into independent factors of variation such as contexts and speaking styles. In a speech domain, we assume the context has a similar property with phonemes or subwords, which are represented with a limited set of discrete symbols. Therefore, to capture the context without any supervision, we use a generative model named a vector quantized variational autoencoder (VQ-VAE) \cite{van2017neural} to extract the discrete symbols. There are several distinctions between a VQ-VAE with a normal autoencoder \cite{vincent2008extracting} and a normal variational autoencoder (VAE) \cite{kingma2014adam}. The VQ-VAE encoder maps the input features to a limited number of discrete latent variables, and a standard VAE encoder maps the input features into continuous latent variables. Therefore, a VQ-VAE encoder has many-to-one mappings due to restricting the representation to the nearest codebook vector, and the standard VAE encoder has one-to-one mapping between the input and latent variables. 
	
	\begin{figure}[]
		\centering
		\includegraphics[width=0.8\linewidth]{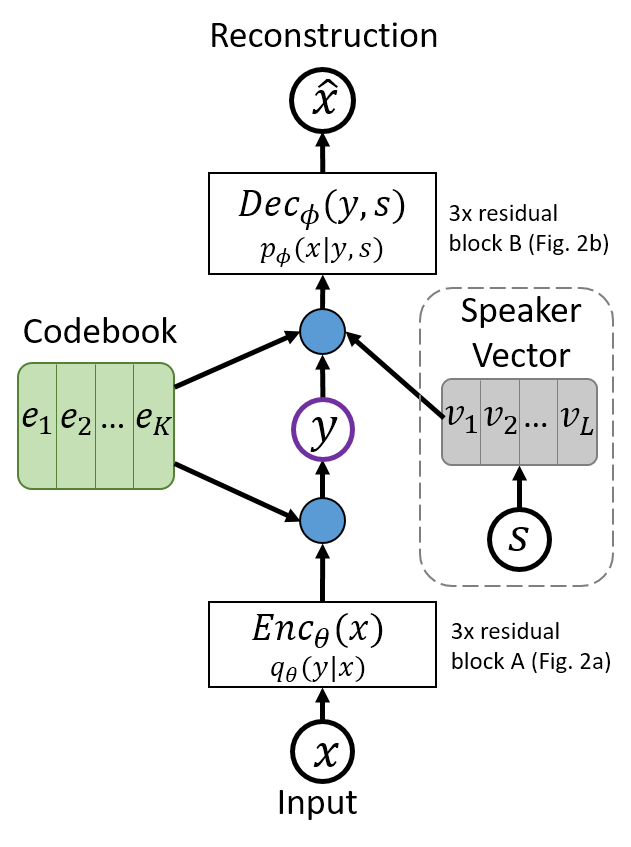}
		\caption{VQ-VAE for unsupervised unit discovery consists of several parts: encoder $\text{Enc}^{VQ}_\theta(x) = q_\theta(y|x)$, decoder $\text{Dec}^{VQ}_\phi(y, s) = p_\phi(x|y,s)$, codebooks $\mbf E =[e_1,..,e_K]$, and (optional) speaker embedding $\mbf V = [v_1,..,v_L]$.}
		\label{fig:vqvae}
	\end{figure}
	
	\begin{figure}[]
		\centering
		\includegraphics[width=0.9\linewidth]{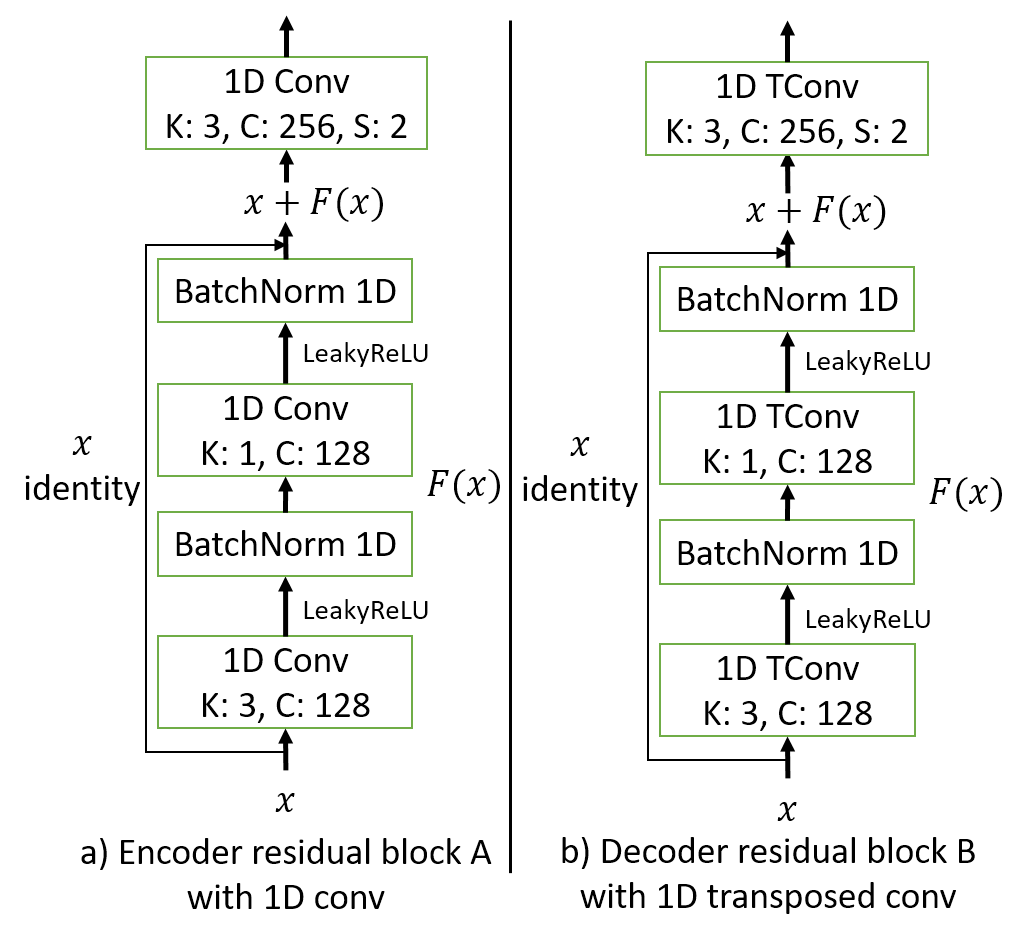}
		\caption{Building block inside VQ-VAE encoder and decoder: a) Encoder residual block and 1D convolution with stride 2 to downsample input sequence length; b) Decoder residual block and 1D transposed convolution with stride 2 to upsample codebook back to original input length.}
		\label{fig:vqvae_resblock}
	\end{figure}

	We illustrate the VQ-VAE model in Fig.~\ref{fig:vqvae} and define $\mathbf{E} = [e_1,..,e_K] \in \R^{K \times D_e}$ as a collection of codebook vectors and $\mathbf{V} = [v_1,..,v_L] \in \R^{L \times D_v}$. During the encoding step, input $x$ is such speech features as MFCC or mel-spectrogram and input $x$'s speaker identity is denoted by $s \in \{1,..,L\}$. In Fig.~\ref{fig:vqvae_resblock}, we show the details for the residual block inside the encoder and decoder modules. Encoder $q_\theta(y|x)$ generates discrete latent variable $y \in \{1,..K\}$ ($y$ can also be represented as a one-hot vector). To transform a continuous representation into a discrete random variable, the encoder first produces intermediate continuous representation ${z} \in \R^{D_e}$. Later, we find which codebook has a minimum distance between ${z}$ and a vector in $\mathbf{E}$. Mathematically, we formulate the operation:
	\begin{eqnarray}
	q_{\theta}(y=c|x) &=& \begin{cases}
	1 \quad \text{if } \, c=\argmin_{i} \text{Dist}( {z}, e_i )\\
	0 \quad \text{else }
	\end{cases} \\
	e_c &=& \mathbb{E}_{q_\theta (y|c)} [\mbf E] \\ &=& \sum_{i=1}^{K} q_\theta(y=i | x) \, e_i.
	\end{eqnarray} where $\text{Dist}(\cdot, \cdot): \R^{D_e} \times \R^{D_e} \rightarrow \R$ is a function to calculate the distance between two vectors. In this paper, we define $\text{Dist}(a, b) = \| a - b \|_2$ as the L2-norm distance.

	After we find closest codebook index $c \in \{1,..,K\}$, we substitute intermediate variable ${z}$ with corresponding codebook vector $e_c$. To reconstruct the input data, decoder $p_\phi(x|y, s)$ reads codebook vector $e_c$ and speaker embedding $v_s$ and generates reconstruction $\hat{x}$. 
	
	The following is the learning objective for VQ-VAE:
	\begin{eqnarray}
	\mathcal{L}_{VQ} = -\log p_\phi(x|y, s) + \gamma \| {z} - \text{sg}(e_c) \|_2^2, \label{eq:loss_vq}
	\end{eqnarray} where function $\text{sg}(\cdot)$ stops the gradient, defined as:
	\begin{eqnarray}
	x &=& sg(x) \\
	\frac{\partial \,\text{sg}(x)}{\partial \,x} &=& 0.
	\end{eqnarray}
	The first term is a negative log-likelihood to measure the reconstruction loss between original input $x$ and reconstruction $\hat{x}$ to optimize encoder parameters $\theta$ and decoder parameters $\phi$. The second term minimizes the distance between intermediate representation ${z}$ and nearest codebook $e_c$, but the gradient is only back-propagated into encoder parameters $\theta$ as commitment loss. Such commitment loss can be scaled with additional hyperparameters $\gamma$. To update the codebook vectors, we use an exponential moving average (EMA) \cite{kaiser2018fast}. With an EMA update rule for training codebook $\mbf E$, the model has a more stable result during the training process and avoids the posterior collapse issue \cite{roy2018towards}.

	\section{Sequence-to-Sequence from Speech to Codebook}
	\label{sec:seq2seq}
\begin{figure}[]
		\centering
		\includegraphics[width=1.0\linewidth]{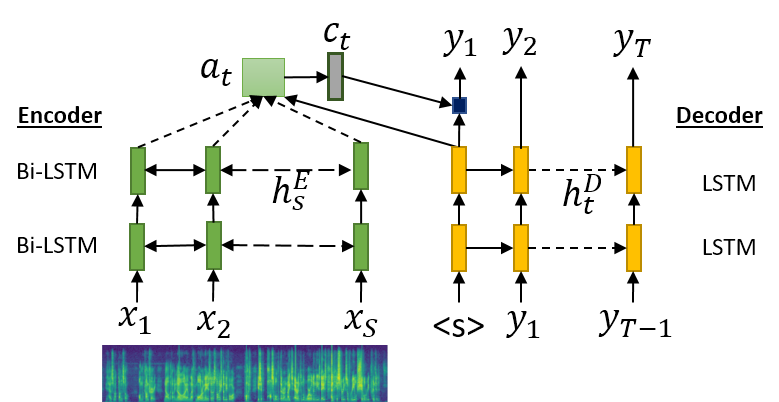}
		\caption{Sequence-to-sequence model with attention mechanism. Here encoder input is speech features $\mbf X = [x_1, .., x_S]$, and decoder predicts codebook index $y_t$ for each time-step.}
		\label{fig:seq2seq}
	\end{figure}

	Our speech-to-speech translation model is built based on an attention sequence-to-sequence (seq2seq) framework \cite{bahdanau2014neural, sutskever2014sequence}. Assume paired source sequence $\mbf X = [x_1, ..., x_S]$ and target sequence $\mbf Y = [y_1, ..., y_T]$.  A sequence-to-sequence model directly learns mapping $P_\psi(\mbf X|\mbf Y)$, parameterized by $\psi$ parameters.
	In this paper, we specify $\mbf X \in \R^{S \times D_s}$ to represent such speech features as MFCC or mel-spectrogram and $\mbf Y = [y_1, ..., y_T] \in \{1,..,K\}$ to represent codebook $\mbf E$ indices. Fig.~\ref{fig:seq2seq} illustrates a seq2seq model with an attention mechanism.
	Inside a seq2seq model, there are three different components: 
	\begin{enumerate}
		\item The encoder module reads all the sequence speech features and represents them with $h^E = [h_1^E, .., h_S^E] \in \R^{S \times M}$ where $h^E=\text{Enc}^{S2S}_{\psi}(\mbf X)$.
		\item The attention module assists the decoder to find which part of the encoder contains related information for current decoding state \cite{bahdanau2014neural}. Given decoder state $h_t^D \in \R^{n}$, the attention modules generate attention $a_t \in \R^S$ and context $c_t \in \R^{N}$: 
		\begin{eqnarray}
		a_t[s] &=& \frac{\exp{\left(Score(h_s^E, h_t^D)\right)}}{\sum_{s=1}^{S} \exp{\left(Score(h_s^E, h_t^D)\right)}} \\
		c_t &=& \sum_{s=1}^{S} a_t[s] \,h_s^E,
		\end{eqnarray} where function $Score(\cdot, \cdot): \R^{M} \times \R^{N} \rightarrow \R$ predicts the relevancy value between the encoder and decoder states. Many $Score$ functions exist, including dot-product \cite{luong-etal-2015-effective}, MLP \cite{bahdanau2014neural} or modified MLP with history \cite{tjandra2018multi}.
		\item The decoder module predicts class probability \\
		$p_t = \text{Dec}^{S2S}_\psi(y_t | c_t, \mbf Y_{<t}, h_t^D)$ over $K$ different classes (depending on codebook $\mbf E$ size) given context $c_t$, previous information $\mbf Y_{< t}$, and current decoder state $h_t^D$.
	\end{enumerate} 
	
	To train a seq2seq model, the most common objective is to minimize the negative log-likelihood over the correct class: 
	\begin{eqnarray}
	\L_{NLL} = -\frac{1}{T} \sum_{t=1}^{T} \sum_{k=1}^{K} \mathbbm{1}(y_t=k) * \log p_t[y=k] \label{eq:loss_nll},
	\end{eqnarray} where $p_t[y = k]$ is the predicted probability on the $k$-th class and time-step $t$.
	
	\section{Codebook Inverter}
	\label{sec:inverter}
	A codebook inverter is an module that synthesizes the corresponding speech utterance from a sequence of the codebook index. Its input, which is a sequence of codebook embedding, is $[\mbf E[y_1],..,\mbf E[y_{T_Y}]]$, and the output target is a sequence of speech representation (e.g., linear magnitude spectrogram) $\mbf X^{R} = [\mbf X^R_1, .., \mbf X^R_{T_{x}}]$. 
	
\begin{figure}[h]
		\centering
		\includegraphics[width=0.9\linewidth]{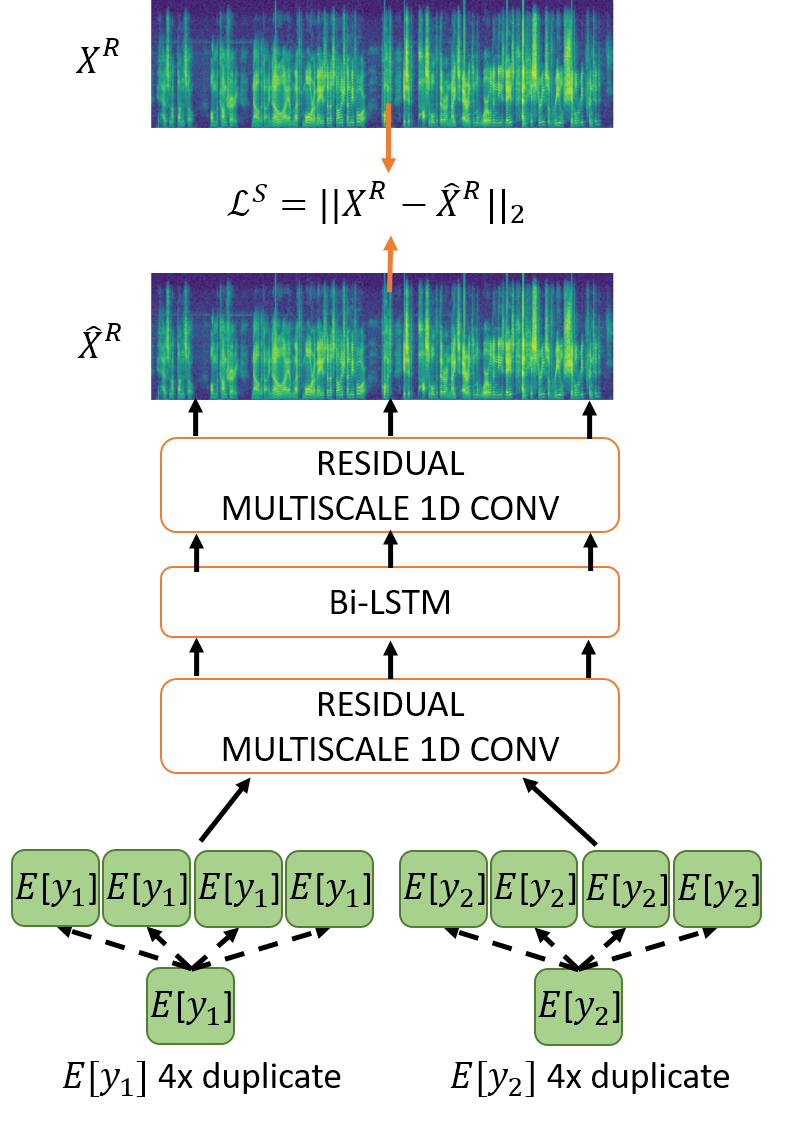}
		\caption{Codebook inverter: given codebook sequence $[\mbf E[y_1], .., \mbf  E[y_{T_Y}]]$, we predict corresponding linear magnitude spectrogram $\hat{\mbf X}^R = [x^R_1, .. x^R_{T_X}]$. If the lengths between $T_Y$ and $T_X$ are different, we consecutively duplicate each codebook by $r$-times.}
		\label{fig:codeinverter}
	\end{figure}

\begin{figure}
		\centering
		\includegraphics[width=0.9\linewidth]{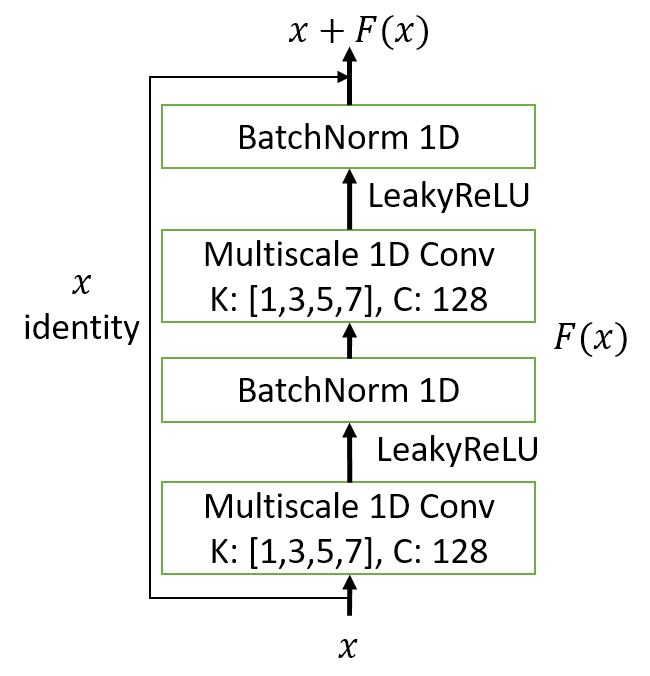}
		\caption{Residual 1D block combines multiscale 1D convolution with different kernel size and ``SAME'' padding, LeakyReLU \cite{xu2015empirical} activation function, and 
			batch normalization \cite{ioffe2015batch}.}
		\label{fig:res1dblock}
	\end{figure}

	We illustrate our codebook inverter architecture in Fig.~\ref{fig:codeinverter}.
	Our codebook inverter is composed of several residual 1D blocks, followed by stacked bidirectional LSTMs \cite{hochreiter1997long}, and finally another several residual 1D blocks.  Fig.~\ref{fig:res1dblock} shows the details inside the block.
	Under certain circumstances, codebook sequence length $T_Y$ might be shorter than $T_X$ because VQ-VAE encoder $q_\theta(y|x)$ has convolution with a stride larger than 1. Therefore, to align the codebook sequence with the speech representation target sequence, we duplicate each codebook $E[y_t]$ into $r$ copies side-by-side where $r= T_X / T_Y$.
	To train a codebook inverter, we set the objective function:
	\begin{eqnarray}
	\L_{INV} =  \| \mbf X^R - \hat{\mbf X}^R \|_{2} \label{eq:loss_inv}
	\end{eqnarray} to minimize the L2-norm between predicted spectrogram $\hat{\mbf X}^R = \text{Inv}_{\rho}([E_[y_1], ..., E[y_{T_Y}]])$ and groundtruth spectrogram $\mbf X^R$. We defined $\text{Inv}_{\rho}$ as the inverter parameterized by $\rho$. In the inference stage, we used Griffin-Lim \cite{griffin1984signal} to reconstruct the phase from the spectrogram and applied an inverse short-term Fourier transform (STFT) to invert it into a speech waveform.

	\section{Training and Inference}
	\begin{figure*}[]
		\centering
		\includegraphics[width=1.0\linewidth]{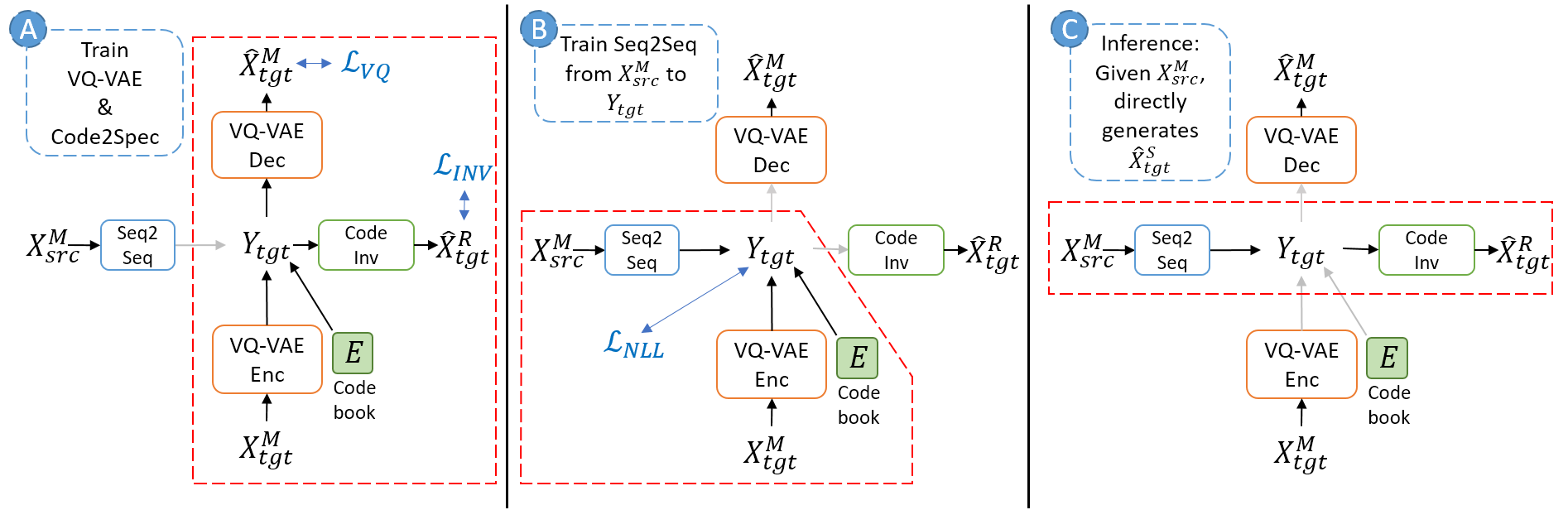}
		\caption{a) Train VQ-VAE to represent  continuous MFCC vectors with codebook sequence and train codebook inverter to generate a linear magnitude spectrogram based on generated codebook sequence; b) Train a seq2seq model from source language MFCC to target language codebook. c) In inference stage, seq2seq model takes source language MFCC and predicts codebook sequences, and then codebook inverter generates target language speech representation.}
		\label{fig:step_by_step}
	\end{figure*}
	In this section, we explain our proposed method in detail and step-by-step. To train our proposed model, we setup three different modules: VQ-VAE (Section~\ref{sec:vqvae}), a speech-to-codebook seq2seq (Section~\ref{sec:seq2seq}), and a codebook inverter (Section~\ref{sec:inverter}). Fig.~\ref{fig:step_by_step} shows which modules are trained in each step. Initially, we defined $\{\mbf X^M_{src}, \mbf X^M_{tgt}\}$ as paired parallel speech, $\mbf X^M_{src}$ is the MFCC features from the source language, and $\mbf X^M_{tgt}$ is the MFCC features from the target language. $\mbf Y_{tgt}$ is the codebook sequences generated by VQ-VAE encoder $\text{Enc}_\theta(x)$ given $\mbf X_{tgt}^M$ as the input. $\hat{\mbf X}^R_{tgt}$ is the predicted linear spectrogram of the target language. $\L_{VQ}, \L_{INV}, and \L_{NLL}$ are calculated by the formula in Eqs.~\ref{eq:loss_vq}, \ref{eq:loss_inv}, and \ref{eq:loss_nll}.
	\begin{enumerate}
		\item First, we trained the VQ-VAE model on target language MFCC $\mbf X_{tgt}^M$. We also trained the codebook inverter to predict corresponding linear spectrogram $\mbf X^{R}_{tgt}$. 
		\item Second, we trained the seq2seq model from the source language speech to the target language codebook. Given a paired parallel MFCC from source and target languages $\{\mbf X_{src}^{M}, \mbf X_{tgt}^M\}$, we extracted codebook sequence $\mbf Y_{tgt} = \text{Enc}^{VQ}_{\theta}(\mbf X_{tgt}^M)$ from the VQ-VAE encoder. Later, we trained the seq2seq translation model to predict $ \hat{\mbf Y}_{tgt} = \text{Seq2Seq}(\mbf X_{src}^{M})$ and minimize loss $\L_{NLL}$ between $\mbf X_{src}^{M}$ and $X_{src}^{M}$.
		
		\item In the inference step, given source language speech $\mbf{X}_{src}^{M}$, we decoded a target language codebook index sequence $\hat{\mbf{Y}}_{tgt} = \text{Seq2Seq}_{\psi}(\mbf{X}_{src}^{M})$ and synthesized it into target language speech $\hat{\mbf X}_{tgt}^{R} = \text{Inverter}(\hat{\mbf Y}_{tgt})$.
	\end{enumerate}

	\section{Experimental Setup}
	
	\subsection{Dataset}
	In this paper, we ran our experiment based on the Basic Travel Expression (BTEC) corpus \cite{kikui2003creating, kikui2006comparative} that has several language pairs . We chose two tasks: French-to-English and Japanese-to-English. For both language pairs, we used the BTEC1 set that consisted of 162,318 training sentences and 510 test sentences. Since the speech utterances for the sentences are unavailable, we generated sentences with Google text-to-speech API for all languages pairs. Even though the lack of natural speech dataset in this paper, VQ-VAE and codebook inverter can be applied and has shown a great performance on multispeaker natural speech \cite{andros2019zero, cho2019zero}. Some papers \cite{tjandra2017listening, tjandra2018machine, tjandra2019stestimator} also show the performance improvement from the synthetic dataset can be carried over to the real dataset.
	
	\subsection{Speech Feature Extraction}
	For the source language and target language speech utterances, we represented the speech utterances with mel-frequency cepstral coefficients (MFCCs) with 13 dimensions $+ \Delta + \Delta^2$ (total 39 dimensions). For the target language speech utterances, we also generated a linear magnitude spectrogram with 1025 dimensions for the codebook inverter (Section~\ref{sec:inverter}) training target. For each frame, we extracted the MFCCs and the linear magnitude spectrogram with a 25-millisecond-sized window and 10-millisecond time-steps. We extracted both the MFCC and the linear magnitude spectrogram with Librosa \cite{mcfee2015librosa} library.
	
	\subsection{Evaluation}
	For an objective evaluation of the target speech utterances, currently there is no standard method can be used to measure translation quality directly on the speech utterances. Therefore, we utilized a pre-trained ASR on the English BTEC dataset and the generated transcription for our evaluation. For the ASR architecture, the encoder module has three stacked Bi-LSTMs with 512 hidden units, and the decoder has one LSTM with 512 hidden units. For the attention module, we utilized MLP attention with multiscale location history \cite{tjandra2018multi}. For the output unit, we used a word-level token from the English transcription. Because there is a performance gap between the ASR and the ground truth cause by imperfect transcription, we assume the metric (calculated based on the ASR transcription) is the lowerbound for the related translation model. We utilized two metrics to evaluate the translation performance from the transcribed text: BLEU scores \cite{papineni2002bleu} and METEOR \cite{banerjee-lavie-2005-meteor} with a Multeval toolkit \cite{clark2011better}. Our pre-trained ASR model resulted in a 2.84\% WER, a 94.9 BLEU, and a 69.1 METEOR on English speech utterances from the BTEC test set, and we set those scores as the groundtruth topline scores.
	
	\section{Results and Discussion}
	In this section, we present our experimental result and followed by the discussion. 
	
	\subsection{Baseline}
	For the baseline translation task, we modified the Tacotron \cite{wang2017tacotron} model by changing the source input from a one-hot character embedding into a continuous vector. Basically, we changed the embedding layer in the encoder layer with a linear projection layer. Therefore, this model directly translated the source language MFCC to a target language mel-spectrogram. However, this approach did not converge at all and produced no audible s peech. \cite{jia2019direct} also observed a similar result with a similar scenario. 
	
	\subsection{Topline with Cascade ASR-TTS}
	In this paper, we set the topline performance by using the cascade of ASR and TTS system. First, we train the ASR system by using the source language MFCC as the input and target language character transcription. Second, we train a TTS based on Tacotron \cite{wang2017tacotron} to generate a speech from the target language characters to the target language speech representation.
	
	\subsection{French-to-English by Speech to Codebook}
	Table~\ref{tab:fr2en} shows our experimental result on various hyperparameters across different codebook sizes and time-reductions. We tried several hyperparameters, including codebook size and time-reduction factor. Our best performance was produced by codebook of 64 and a time-reduction factor of 12 with a score of 25.0 BLEU and 23.2 METEOR.
	\begin{table}[]
		\caption{Our experiment results based on BTEC French-English speech-to-speech translation:}
		\label{tab:fr2en}
		\centering
		\begin{tabular}{|c|c|c|c|}
			\hline
			\multicolumn{2}{|c|}{\textbf{Model (FR-EN)}} & \textbf{BLEU} & \textbf{METEOR} \\ \hline \hline
			\multicolumn{2}{|c|}{\textbf{\begin{tabular}[c]{@{}c@{}}Baseline \\ Tacotron with MFCC input\end{tabular}}} & - & - \\ \hline
			\multicolumn{4}{|c|}{\textbf{Proposed Speech2Code}} \\ \hline
			Codebook & \begin{tabular}[c]{@{}c@{}}Time \\ Reduction\end{tabular} &  &  \\ \hline
			32 & 4 & 19.4 & 19.1 \\ \hline
			32 & 8 & 23.8 & 22.2 \\ \hline
			32 & 12 & 23.2 & 22.1 \\ \hline
			64 & 4 & 16.1 & 16.9 \\ \hline
			64 & 8 & 24.4 & 22.9 \\ \hline
			64 & 12 & 25.0 & 23.2 \\ \hline
			128 & 4 & 16.9 & 17.4 \\ \hline
			128 & 8 & 23.3 & 22.1 \\ \hline
			128 & 12 & 24.2 & 21.9 \\ \hline
			\multicolumn{2}{|c|}{\textbf{\begin{tabular}[c]{@{}c@{}}Topline \\ (Cascade ASR -\textgreater TTS)\end{tabular}}} & 47.4 & 41.2 \\ \hline
		\end{tabular}
	\end{table}

	\subsection{Japanese-to-English by Speech to Codebook}
	Table~\ref{tab:ja2en} shows our experimental result on various hyperparameters across different codebook sizes and time-reductions. We tried several hyperparameters, including codebook size and time-reduction factor. Our best performance was produced by a codebook of 128 and a time-reduction factor 8 with a score of 15.3 BLEU and 15.3 METEOR.
	\begin{table}[]
		\caption{Our experiment results based on BTEC Japanese-English speech-to-speech translation.}
		\label{tab:ja2en}
		\centering
		\begin{tabular}{|c|c|c|c|}
			\hline
			\hline
			\multicolumn{2}{|c|}{\textbf{Model (JA-EN)}}                                                                      & \textbf{BLEU} & \textbf{METEOR} \\ \hline
			\multicolumn{2}{|c|}{\textbf{\begin{tabular}[c]{@{}c@{}}Baseline\\ Tacotron with MFCC source\end{tabular}}}       & -             & -               \\ \hline
			\multicolumn{4}{|c|}{\textbf{Proposed Speech2Code}}                                                                                                 \\ \hline
			Codebook                                              & \begin{tabular}[c]{@{}c@{}}Time \\ Reduction\end{tabular} &               &                 \\ \hline
			32                                                    & 4                                                         & 14.8          & 15              \\ \hline
			32                                                    & 8                                                         & 14.2          & 15.6            \\ \hline
			32                                                    & 12                                                        & 16            & 16              \\ \hline
			64                                                    & 4                                                         & 10.8          & 12.1            \\ \hline
			64                                                    & 8                                                         & 14.2          & 14.7            \\ \hline
			64                                                    & 12                                                        & 14.7          & 14.8            \\ \hline
			128                                                   & 4                                                         & 11.9          & 13.5            \\ \hline
			128                                                   & 8                                                         & 15.3          & 15.3            \\ \hline
			128                                                   & 12                                                        & 14.9          & 14.5            \\ \hline
			\multicolumn{2}{|c|}{\textbf{\begin{tabular}[c]{@{}c@{}}Topline \\ (Cascade ASR -\textgreater TTS)\end{tabular}}} & 37.4          & 32.8            \\ \hline
		\end{tabular}
	\end{table}
	\subsection{Transcription Example and Discussion}
	In Table \ref{tab:result_transcript}, we provide some transcriptions example from the ground-truth, our proposed speech-to-code and topline cascade ASR-TTS models. In the first result, all models translation contains similar meaning with the ground truth. In the second result, all models still maintain a similar semantic with the ground-truth. However, compared to the topline, the speech-to-code does not produce the additional translation for ``as soon as he comes in''. In the third result, our proposed method can only translate the beginning of the sentence correctly and produce incorrect result in the latter part. From the transcription result, the missing part and arbitrary transcription in the latter half might be interesting to be investigated in the future.
	
	For further information and translation samples, our reader could refer to:\\
	\href{https://sp2code-translation-v1.netlify.com/}{\url{https://sp2code-translation-v1.netlify.com/}}.
	
	\begin{table}[]
		\caption{Transcription example between the ground truth, our proposed Speech2Code, and topline (Cascade ASR-TTS) model.}
		\label{tab:result_transcript}
		\begin{tabular}{|c|l|}
			\hline
			\textbf{Model} & \multicolumn{1}{c|}{\textbf{Transcription Result}} \\ \hline \hline
			Groundtruth & how long are you going to stay \\ \hline 
			\begin{tabular}[c]{@{}c@{}}Speech2Code \\ FR-EN\end{tabular} & how long are you going to stay \\ \hline
			\begin{tabular}[c]{@{}c@{}}Speech2Code \\ JA-EN\end{tabular} & how long will it take \\ \hline
			Topline FR-EN & how long are you staying \\ \hline
			Topline JA-EN & how long are you staying \\ \hline \hline
			Groundtruth & please tell him to call me as soon as he comes in \\ \hline
			\begin{tabular}[c]{@{}c@{}}Speech2Code \\ FR-EN\end{tabular} & please tell him to call me back \\ \hline
			\begin{tabular}[c]{@{}c@{}}Speech2Code \\ JA-EN\end{tabular} & please tell him that i called \\ \hline
			Topline FR-EN & please tell her to call me and check it \\ \hline
			Topline JA-EN & please ask him to call me as soon as possible \\ \hline\hline
			Groundtruth & i would like a balcony seat please \\ \hline
			\begin{tabular}[c]{@{}c@{}}Speech2Code \\ FR-EN\end{tabular} & i would like to have this film please \\ \hline
			\begin{tabular}[c]{@{}c@{}}Speech2Code \\ JA-EN\end{tabular} & i would like a seat near the seat \\ \hline
			Topline FR-EN & i would like a balcony seat please \\ \hline
			Topline JA-EN & i would like a balcony seat \\ \hline
		\end{tabular}
	\end{table}
	
	\section{Conclusion}
	In this paper, we proposed a novel approach for training a speech-to-speech translation between two languages without any transcription. First, we trained a discrete quantized autoencoder to generate a discrete representation from the target speech features. Second, we trained a sequence-to-sequence model to predict the codebook sequence given the source speech representation. This method is  applicable to any type of language, with or without a written form because the target speech representations are trained and generated unsupervisedly. Based on our experiment result, our model can perform a direct speech-to-speech translation on French-English and Japanese-English. 
	\section{Acknowledgments}
	Part of this work was supported by JSPS KAKENHI Grant Numbers JP17H06101 and JP17K00237.

	\bibliographystyle{IEEEbib}
	\bibliography{strings,refs}
	
\end{document}